# Hybrid LQG-Neural Controller for Inverted Pendulum System

E.S. Sazonov
Department of Electrical and Computer Engineering
Clarkson University
Potsdam, NY 13699-5720 USA

P. Klinkhachorn and R. L. Klein
Lane Dept. of Computer Science and Electrical Engineering,
West Virginia University,
Morgantown, WV 26506-6109 USA

*Abstract* — The paper presents a hybrid system controller, incorporating a neural and an LQG controller. The neural controller has been optimized by genetic algorithms directly on the inverted pendulum system. The failure-free optimization process stipulated a relatively small region of the asymptotic stability of the neural controller, which is concentrated around the regulation point. The presented hybrid controller combines benefits of a genetically optimized neural controller and an LQG controller in a single system controller. High quality of the regulation process is achieved through utilization of the neural controller, while stability of the system during transient processes and a wide range of operation are assured through application of the LQG controller. The hybrid controller has been validated by applying it to a simulation model of an inherently unstable system – inverted pendulum.

## I. INRODUCTION

The traditional approach to building system controllers requires a prior model of the system. The quality of the model, that is, loss of precision from linearization and/or uncertainties in the system's parameters negatively influence the quality of the resulting control.

At the same time, methods of soft computing such as neural networks or fuzzy logic possess non-linear mapping capabilities, do not require an analytical model and can deal with uncertainties in the system's parameters. Combined with the evolutionary learning (such as genetic algorithms) these methods are capable of producing near-optimal controllers for a given control task. For example, genetic algorithms have been used to produce parameters of an optimized system controller such as the architecture and/or weights of a neural network controller [1,2], rules and/or membership functions of a fuzzy controller [3,4], and to obtain model equations [5], etc.

The disadvantage of the Genetic Algorithms (GA) is that the process routinely produces solutions (parameter sets of a controller) that may render the controlled system unstable.

A failure-free optimization method employing GA and a neural controller has been described in [6]. The suggested method applies evolutionary learning to a neural controller in a subspace around the regulation point to ensure a failure-free optimization process. Thus, due to the nature of the failure-free learning methodology, the optimized neural controller is capable of controlling the system in a relatively small region of state space, which may be a limiting factor for some practical applications of the optimized neural controller.

This paper presents a hybrid controller that combines the benefits of an optimized neural controller and an LQG controller in a single system controller. The high quality of regulation process is ensured by application of the optimized neural controller, while the wide range of operation and stability of transient processes is provided by the LQG controller.

## II. TEST BED

A numerical model of an Inverted Pendulum (IP) served as the test bed for the development of the proposed hybrid controller. Utilization of a model instead of an actual system allowed expediting and simplifying the experimentation process. The performance and accuracy of the model was verified during the design of the LQG controller [7].

The IP system consists of a cart sliding on a rail and a rod pivoted to the cart and free to rotate about an axis perpendicular to the direction of motion of the cart. The system is equipped with two sensors measuring cart position and rod angle, and a DC motor providing actuation control. The numerical model of the IP system not only simulates the dynamics of IP motion, including saturations on the state variables of cart position and rod angle, but also accounts for major non-linearities of the system, including the dead zone and saturation of the DC motor input voltage and force it can produce. Additionally, the model incorporates such parameters as sensor offsets, discretization errors and measurement noise. More details of the model, including corresponding modeling equations can be found in [7].

## III. EXPERIMENTAL SETUP

This paper describes a hybrid controller that utilizes a neural and an LQG controller. The block diagram of the hybrid controller is presented in Fig. 1. The numerical model described in section II simulates an inverted pendulum system. A linearized model of the IP dynamics was adopted for the purpose of designing the LQG controller:

$$\begin{cases} (M+m)\ddot{p}(t) + m\dfrac{l}{2}\ddot{\Theta}(t) = C_V V(t) - (C_p + \beta)\dot{p}(t) \\ -m\dfrac{l}{2}\ddot{\Theta}(t) = m\dot{p}(t) \end{cases} \quad (1)$$

where $M$ and $m$ are rod and cart masses respectively, $l$ is the rod length, $p(t)$ is the cart position with respect to the center of the rail, $\Theta(t)$ is the rod angle with respect to the vertical, $C_V$ is the motor torque constant, $V(t)$ is the voltage supplied to the electric motor, $C_P$ and $\beta$ are the coefficients reflecting the dynamic and static friction in the coupling between the motor shaft and the rail. A detailed description of the LGQ controller can be found in [7].

The neural controller is a multi-layer perceptron (neural network) composed of the neurons with the sigmoid transfer function. The neural controller has a fixed architecture:

a. Four inputs – cart position in meters (from –0.5 to 0.5 with respect to the center of the rail); cart velocity in meters per second (from -5.0 to 5.0); rod angle in radians (from -0.5 to 0.5 with respect to the vertical) and rod angular velocity in radians per second (from –5.0 to 5.0).
b. Two hidden layers, 4 and 2 neurons each.
c. One output. The neurons used in this neural network could only provide output in the range from 0.0 to 1.0, which is later scaled to the range from 0.0 to 5.0 (motor control voltage).

The neural controller has been subjected to a period of genetic optimization through the SAFE-LEARNING method described in [6]. The optimization goal was to produce a neural controller (LEARNING controller) that has a better steady state performance (expressed in the terms of RMS error of the cart position and rod angle during rod balancing) than the original SAFE controller. Due to the nature of the failure-free learning method, the training process was limited to a closed neighborhood (further denoted as $\Omega_{SAFE}$) surrounding the regulation point. The neural controller never observed state variables exceeding the boundaries of $\Omega_{SAFE}$, and, therefore, cannot be used outside of that region. The optimization process emphasized optimization of the cart position RMS or rod angle RMS through utilization of the weight coefficients $P_W$ (position weight coefficient in meters)

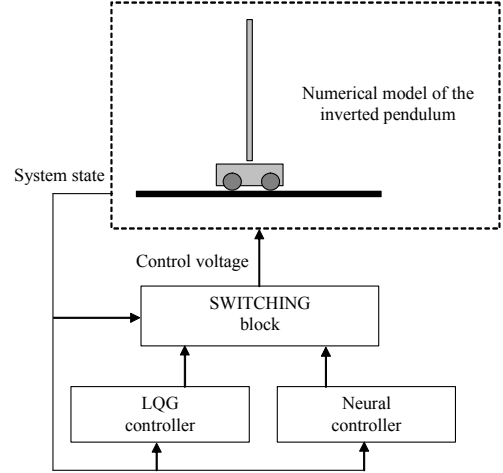

**Fig. 1:** Block-diagram of the hybrid controller.

and $A_W$ (angle weight coefficient in degrees) in the fitness function $F$ of GA.

$$F = \int_0^T \left(\dfrac{P(t)}{P_W}\right)^2 + \left(\dfrac{A(t)}{A_W}\right)^2 dt, \quad (2)$$

where $P(t)$ is cart position in meters, $A(t)$ is rod angle in degrees. Average cart position and rod angle RMS of the inverted pendulum system controlled by the neural controllers optimized with different coefficients $P_W$ and $A_W$ are listed in Table 1. The neural controller optimized with equal importance of the cart position and rod angle was chosen to be used in the hybrid controller.

Also, it is noted that all of the neural controllers optimized by GA produced bang-bang type of control, in contrast to the continuous output of the LQG controller.

The SAFE controller in the SAFE-LEARNING method is a controller that has been validated for performance and stability of operation, though it might not be an optimal controller. The SAFE controller provides a control design, which is ensured to be failure-free even in cases in which the GA optimization process may generate unacceptable solutions. The same LQG controller has been used both as the SAFE controller during optimization and as a part of the hybrid controller.

**Table 1:** The average RMS of cart position and rod angle obtained on the neural controllers optimized with different weight coefficients $P_W$ and $A_W$. Relative improvement in RMS is given in comparison to the LQG controller.

| # | $P_W$, centimeters | $A_W$, degrees | Cart position RMS, centimeters | Rod angle RMS, degrees | Reduction in cart position RMS, % | Reduction in rod angle RMS, % |
|---|---|---|---|---|---|---|
| 1 | 0.5 | 2.0 | 0.6527 | 0.6431 | 53.07 | 9.83 |
| 2 | 0.5 | 1.0 | 0.6553 | 0.6131 | 52.89 | 14.04 |
| 3 | 0.5 | 0.5 | 0.692 | 0.5705 | 50.25 | 20.01 |
| 4 | 1.0 | 0.5 | 0.7759 | 0.5048 | 44.22 | 29.22 |
| 5 | 2.0 | 0.5 | 0.8508 | 0.4795 | 38.83 | 32.76 |

The switching block monitors the state of the controlled system and switches control from the LQG controller to the neural controller and back. The suggested principle of operation for the switching block is illustrated in Fig. 2. The system starts at some initial state $S_{INIT}$, with the LQG controller in control of the system. After a transient process, the system state becomes sufficiently close to the regulation point $S_0$. Subspace $\Omega_N$ defines the region where the current state of system is considered to be close enough to $S_0$, so that the control can be turned over to the neural controller. The neural controller assumes control of the system and continues it until the system state exceeds boundaries of the region of normal operation $\Omega_L$. This event may be the result of changing the reference point of the regulation process. Given such an event, the control is turned over to the LQG controller until the transient process is complete.

The switching block is probably the most important part of the hybrid controller. The quality of the regulation process depends upon timely switching from LQG to neural controller when the current state is within $\Omega_N$. The system's performance will be unacceptable if the switching from the neural to the LQG controller is too late and the LQG controller is not able to recover when the system state transitions outside of $\Omega_L$.

Practical issues related to the implementation of the suggested method include (but are not limited to) a reliable definition of the regions $\Omega_N$ and $\Omega_L$. Region $\Omega_N$ should correspond to the steady state mode of operation for the LQG controller. The switching block should turn control over to the neural controller only during steady state operation, but not during a transient process. Region $\Omega_L$ should correspond to the steady state mode of operation for the neural controller, which may or may not be equal, smaller or larger than $\Omega_N$. Size of the subspace $\Omega_L$ is a result of genetic optimization of a neural controller and will vary depending on the optimization goals. However, region $\Omega_L$ (and $\Omega_N$) should always be a subspace of the region $\Omega_{SAFE}$ in which the neural controller was optimized:

$$\begin{cases} \Omega_L \subset \Omega_{SAFE} \\ \Omega_N \subset \Omega_{SAFE} \end{cases} \qquad (3)$$

Both region $\Omega_N$ and region $\Omega_L$ can be experimentally established by observing balancing on the inverted pendulum system by the LQG and the neural controller, respectively. The actual definition of the regions may be obtained as:
1. A neural network mapping the region of steady state operation.
2. A statistical mapping, such as a clustering technique.
3. An enclosing hypercube or a hypersphere.

The hypercube approach is the simplest but the least accurate of those listed. The hypercube mapping was selected for use in the hybrid controller due to simplicity of implementation. Further development of the hybrid controller will include improved mapping techniques.

The boundaries of the hypercube can be easily obtained by observing the steady state operation of a controller for a sufficiently long period of time. Fig. 3 illustrates histograms of the cart position, cart velocity, rod angle and rod angular velocity for the LQG and for the neural controller during an operation period of 1000 seconds. The hypercube region for switching from the LQG to the neural controller $\Omega_{NHC}$ is specified by the switching limits with respect to the regulation point $S_{REG}$. However, being a crude approximation of the $\Omega_N$, a hypercube may also include states that can only be observed during a transient process. A possible solution to this problem is to monitor the state transitions of the system in time and perform switching from the LQG to the neural controller only after a period of time $T_{SW}$ that system spends in the region $\Omega_{NHC}$. Such a switching mechanism reflects on the properties of the transient processes, which will transition through $\Omega_{NHC}$ in a relatively short time, while a steady state process should remain inside $\Omega_{NHC}$ indefinitely. The hypercube region for switching from the neural to the LQG controller $\Omega_{LHC}$ is defined similarly to $\Omega_{NHC}$. Being an imprecise approximation of $\Omega_L$, $\Omega_{LHC}$ may create situations, where switching from the neural to the LQG controller is performed late, reducing the quality of control. However, as long as $\Omega_{LHC} \subset \Omega_{SAFE}$, the system should remain stable during the transition.

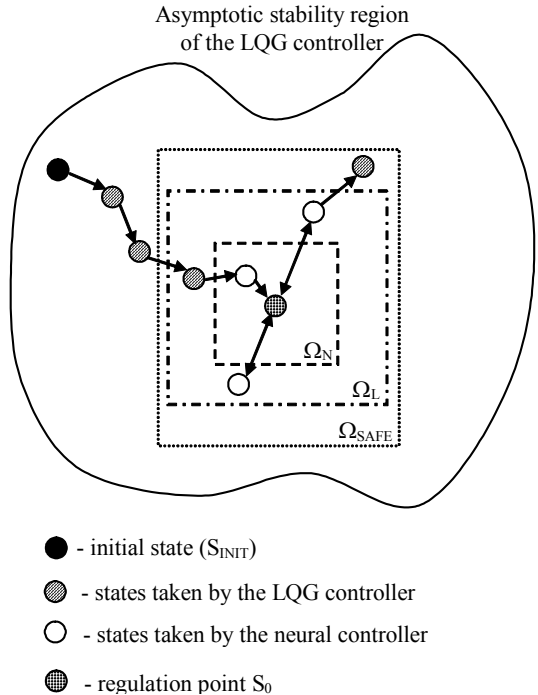

- initial state ($S_{INIT}$)
- states taken by the LQG controller
- states taken by the neural controller
- regulation point $S_0$

**Fig. 2**: A two dimensional example of the switching block operation

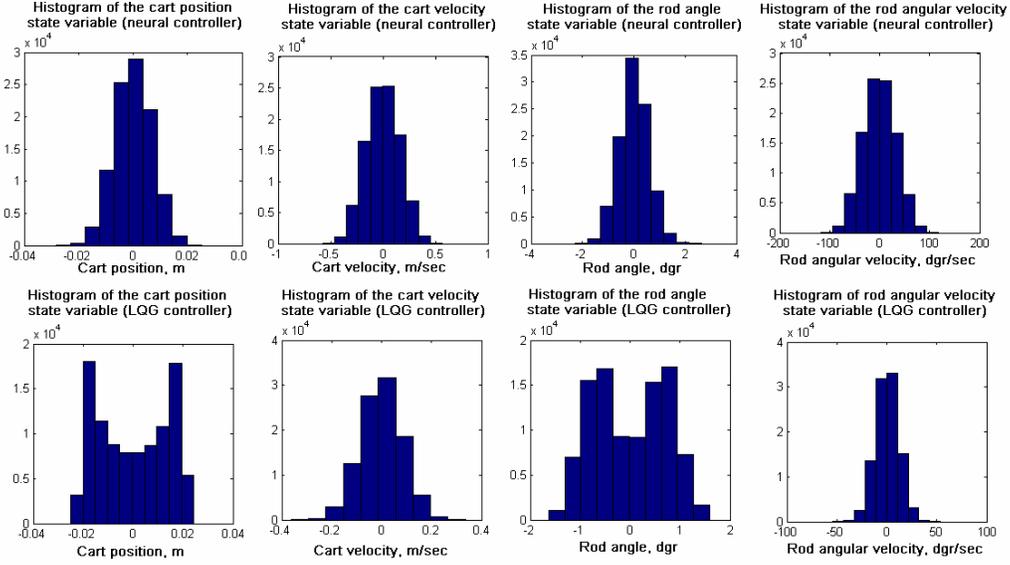

**Fig. 3:** Two-dimensional projections of system variables during 30-second balancing by a neural controller

The flowchart of the switching algorithm is presented in Fig. 4.

## IV. RESULTS

Several experiments were conducted with the different parameters of the reference signals. The hypercube boundaries were established from the histograms shown in Fig. 3. The boundaries were established as a range containing 99% of the observed values. The following numerical experiments were conducted for the LQG and for the hybrid controller:

1. Balancing the pole with the initial conditions close to zero.
2. Balancing from non-zero (cart offset 0.15 m) initial conditions.
3. Tracking a low-frequency (frequency 0.05Hz, amplitude 0.15 m) square wave.
4. Tracking a high-frequency (frequency 0.5Hz, amplitude 0.15 m) square wave.

The average RMS of the cart position and pole angle of the inverted pendulum system obtained during a 100-second run period are listed in Table 2.

As expected, the hybrid controller offered the best improvement in the quality of control for the balancing problem with initial conditions close to zero. The system almost immediately switches control to the neural controller, which takes control for the remaining time.

Similar to the first experiment, the hybrid controller offered significant improvement for the case of non-zero initial conditions. However, for the third experiment, the hybrid controller provided poorer performance than the stand-alone LQG controller. Such an effect may appear due to frequent switching from the LQG controller to the neural controller and back in a system with coarse approximation of $\Omega_N$ and $\Omega_L$. Finally, experiment number four did not demonstrate any improvement over the stand-alone LQG controller. Such a result was expected as the hybrid controller stays switched to the LQG during the transient processes.

Fig. 5 illustrate operation of the hybrid controller for the experiment with the 0.05 Hz square wave.

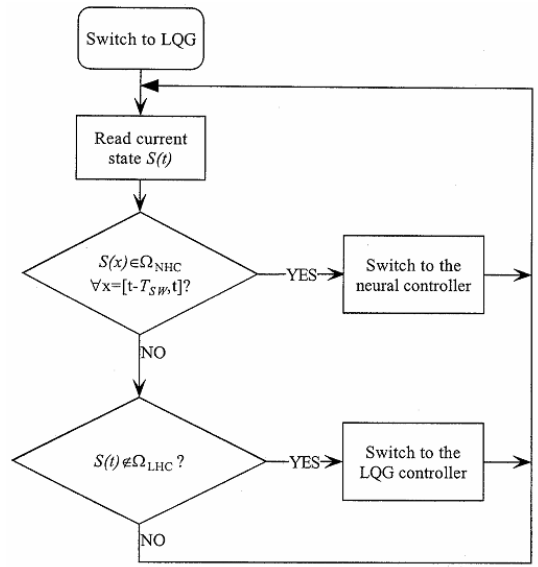

**Fig. 4:** Flowchart of the switching algorithm

Table 2: The average RMS of cart position and rod angle obtained on the stand alone LQG and the hybrid controllers.

| Controller | Parameter | Balancing with zero initial conditions | Balancing with 0.15 m initial cart offset | Tracking 0.05 Hz square wave | Tracking 0.5 Hz square wave |
|---|---|---|---|---|---|
| LQG | Cart position RMS, m | 0.014 | 0.019 | 0.077 | 0.20 |
|  | Rod angle RMS, dgr | 0.72 | 0.75 | 1.55 | 4.03 |
| Hybrid | Cart position RMS, m | 0.0064 | 0.016 | 0.095 | 0.20 |
|  | Rod angle RMS, dgr | 0.56 | 0.62 | 2.29 | 4.03 |

## V. CONCLUSIONS

The hybrid controller, described in this paper, has shown certain advantages over conventional LQG controller:
1. A neural controller, trained directly on the controlled system, accounts for the existing non-linearities and uncertainties of the parameters, improving quality of control.
2. As a part of the hybrid, the LQG controller provides a much wider region of operation than the neural controller alone and ensures stability of the system operation during transient processes.
3. Conducted experiments have shown that the hybrid controller outperforms the stand-alone LQG controller for a variety control tasks.

The coarse switching mechanism used here has demonstrated feasibility of the hybrid controller approach, but lacks robustness. Further research is necessary to completely develop the functioning of the switching block in order to provide the best quality of control and stability of the hybrid controller.

## VI. ACKNOWLEDGEMENTS

The authors wish to acknowledge the support provided for this work by Allegheny Power and the simulation model designed by Dr. Diego DelGobbo.

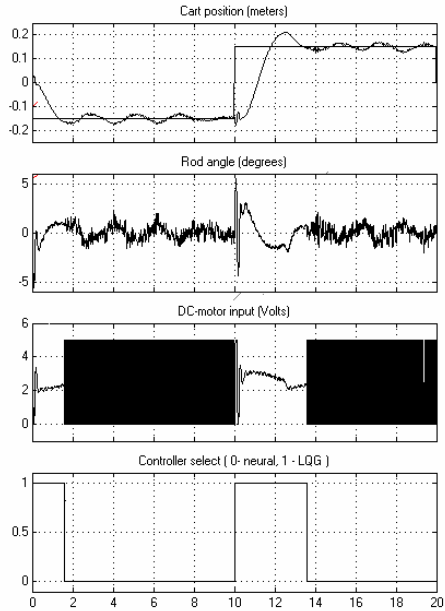

**Fig. 5:** Cart position, rod angle, motor control voltage and controller select signal acquired from the hybrid controller during a balancing experiment with non-zero initial conditions